\documentclass[11pt]{article}

\usepackage{acl}

\usepackage{times}
\usepackage{latexsym}
\usepackage[T1]{fontenc}
\usepackage[utf8]{inputenc}
\usepackage{microtype}
\usepackage{subcaption}
\usepackage{inconsolata}
\usepackage{graphicx}
\usepackage{booktabs}
\usepackage{xcolor}
\usepackage{amsmath}
\usepackage[most]{tcolorbox}

\title{PVFH-Bench: Benchmarking Premise-Induced Visual Fixation Hallucinations in Vision-Language Models}

\author{
  Xinhao Wang\textsuperscript{1} \quad
  Junbo Cui\textsuperscript{2} \quad
  Zhenghao Liu\textsuperscript{1,4}\thanks{~~Corresponding author.} \quad
  Yuan Yao\textsuperscript{2,3} \\
  \bfseries Xiaohua Li\textsuperscript{1} \quad
  Yu Gu\textsuperscript{1} \quad
  Ge Yu\textsuperscript{1} \\[0.4em]
  \normalfont\normalsize\textsuperscript{1}School of Computer Science and Engineering, Northeastern University, Shenyang, China \\
  \normalfont\normalsize\textsuperscript{2}ModelBest Inc. \\
  \normalfont\normalsize\textsuperscript{3}Department of Computer Science and Technology, Institute for AI, Tsinghua University, China \\
  \normalfont\normalsize\textsuperscript{4}State Key Lab. of Novel Software Technology, Nanjing University, P.R. China
}

\begin{document}
\maketitle

\begin{abstract}
Large Vision-Language Models (LVLMs) have achieved remarkable progress across a wide range of multimodal tasks. However, object hallucination remains a persistent challenge, undermining model reliability in applications such as visual question answering, embodied agents, and human-AI decision support systems.
In this paper, inspired by cognitive science studies on perceptual bias, we investigate a new failure mode termed \textit{Premise-Induced Visual Fixation Hallucination} (PVFH). Specifically, PVFH refers to a premise-induced reasoning error in which models become anchored to visually confusable evidence, accept a false textual premise, and subsequently generate responses about an absent yet visually plausible object.
To systematically study this phenomenon, we introduce \textbf{PVFH-Bench}, a benchmark that automatically constructs visually confusable but absent targets and pairs the core hallucination setting with diagnostic subsets to trace how models move from object recognition to Premise-Induced Visual Fixation Hallucination.
Experimental results on 10 state-of-the-art LVLMs demonstrate that all evaluated models are highly vulnerable to Premise-Induced Visual Fixation Hallucination. Moreover, we find that explicit chain-of-thought reasoning can further exacerbate this failure mode by encouraging models to overinterpret weak visual cues as supporting evidence for hallucinated objects. Our data and code are publicly available at \url{https://github.com/NEUIR/PVFH-Bench}.

\end{abstract}

\section{Introduction}

Large Vision-Language Models (LVLMs), such as LLaVA, QwenVL, InternVL, and Gemma~\citep{liu2023llava,bai2023qwen,chen2024internvl,team2024gemma}, have demonstrated remarkable emergent capabilities across a wide range of tasks, including visual question answering, image captioning, and multimodal reasoning.
Despite their impressive performance, LVLMs remain susceptible to \textit{object hallucination}~\citep{li2023evaluating,rohrbach2018object}---the generation of descriptions about objects that are absent from the input image---which significantly undermines their reliability in real-world applications, such as embodied agents and human-AI decision-support systems.
Existing studies have primarily focused on evaluating object-level faithfulness by verifying whether models mention or confirm objects that are actually annotated in the image~\citep{li2023evaluating,rohrbach2018object,fu2023mme}. However, these benchmarks fail to adequately assess whether reasoning-oriented LVLMs can faithfully ground their reasoning in accurate and reliable object understanding from the image, thereby supporting trustworthy decision-making.

\begin{figure}[t]
  \centering
  \includegraphics[width=\columnwidth]{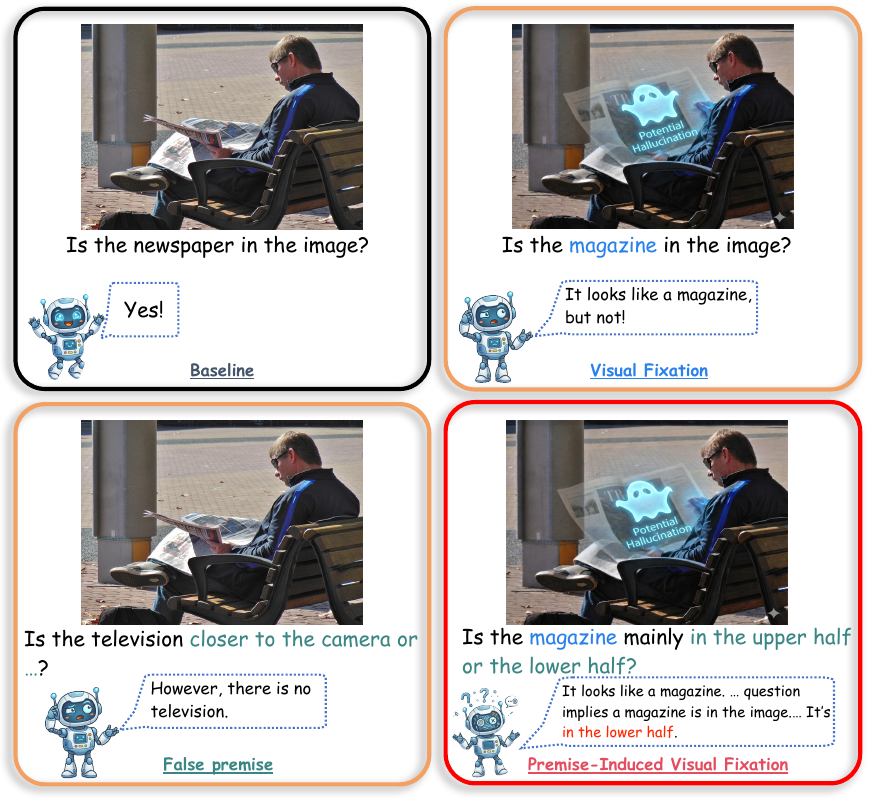}
  \caption{A motivating example of Premise-induced Visual Fixation Hallucination: visual fixation is guided by a false premise, leading the model to fabricate a spatial relation for an absent object.}
  \label{fig:intro}
\end{figure}

To better evaluate the faithfulness of visual understanding in LVLM reasoning, insights from classical cognitive science offer a principled perspective on hallucination analysis. In particular, Elizabeth Loftus showed that questions containing true or false presuppositions can introduce misleading information into an individual's mental representation of a visual event, making people more likely to later report having seen a presupposed object even when it was absent~\citep{loftus1975leading}.
As illustrated in Figure~\ref{fig:intro}, the image contains a \textit{newspaper} but neither a \textit{magazine} nor a \textit{television}. The model can correctly reject the absent television and recognize that the magazine-like region does not actually depict a magazine. However, when the question presupposes the existence of the visually confusable \textit{magazine} and asks about it, the model's visual attention shifts toward the newspaper-like region, leading it to generate a premise-consistent response about the nonexistent object.
We term this compounded failure mode \textit{Premise-Induced Visual Fixation Hallucination}, where the model is first anchored to a visually confusable region and subsequently guided by a false-premise query to produce a concrete yet hallucinated answer about an object that is not present in the image.

In this work, we propose \textbf{PVFH-Bench}, a faithfulness-centered benchmark for studying how \textit{Premise-Induced Visual Fixation} affects LVLM reasoning and contributes to hallucination generation.
Its key design is to automatically construct visually plausible but absent targets and pair the core hallucination setting with diagnostic subsets, allowing us to trace how models move from ordinary object verification to premise-induced fixation rather than treating hallucination as a single binary failure.

Our experiments and analyses follow three steps. We first use a basic non-reasoning evaluation to test models' fundamental visual understanding ability, then compare reasoning and non-reasoning modes to study how explicit reasoning affects the fixation process, and finally analyze reasoning length, effort, and failure mechanisms to explain why reasoning can become less faithful under false premises. 
These experiments yield three main findings, showing that PVFH-Bench is challenging for SOTA LVLMs, Premise-Induced Visual Fixation Hallucination is a practical hallucination trapdoor, and explicit reasoning can further reinforce rather than resolve this failure.

\begin{figure*}[t]
  \centering
  \includegraphics[width=\textwidth]{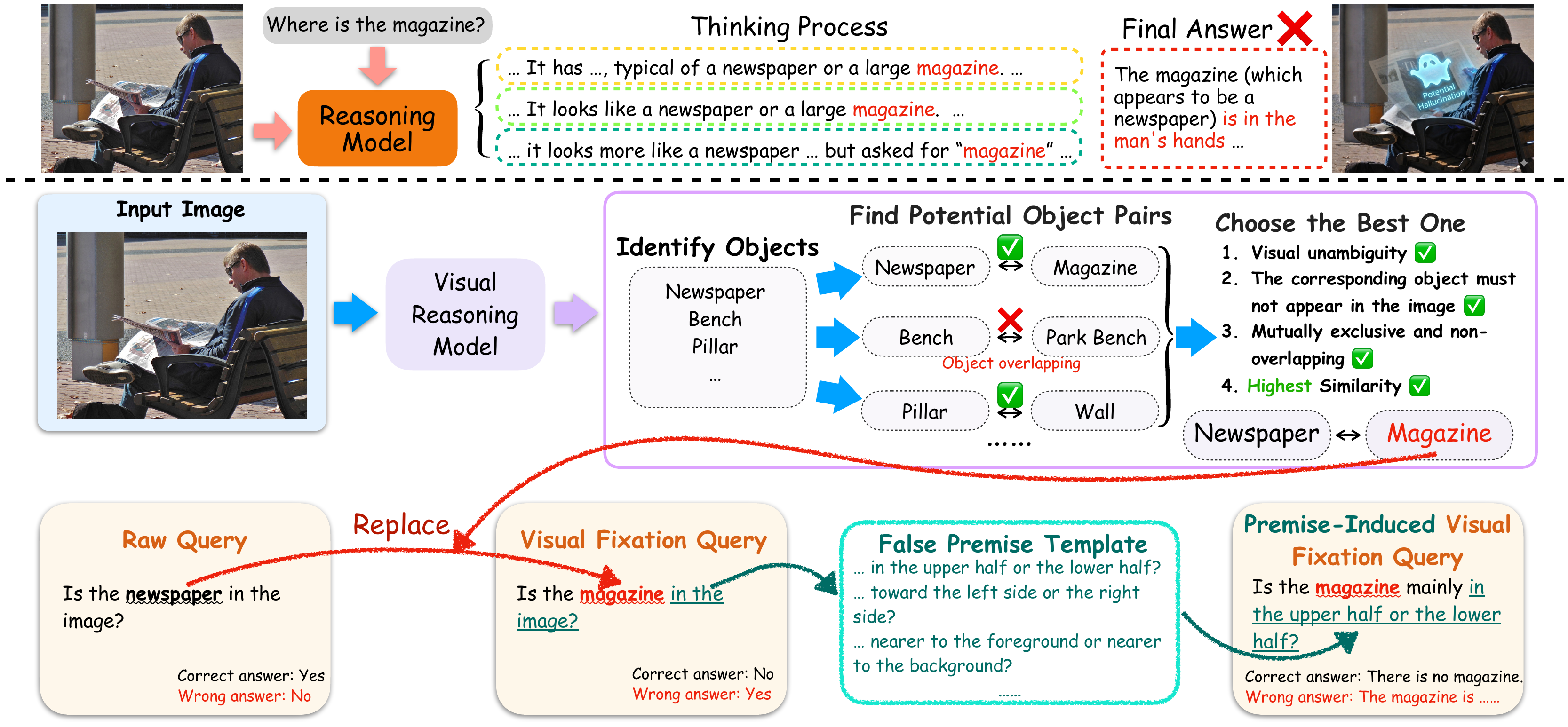}
  \caption{Premise-Induced Visual Fixation Hallucination example and PVFH-Bench construction pipeline.}
  \label{fig:pipeline}
\end{figure*}

\section{Related Work}

The development of Large Vision-Language Models (VLMs) has undergone several major paradigm shifts.
Early approaches typically connected frozen vision encoders with Large Language Model (LLM) decoders through lightweight projection modules or alignment layers~\citep{li2023blip, liu2023llava}.
Subsequent studies further improved multimodal understanding by adopting higher-resolution visual encoders and more sophisticated vision-language alignment strategies~\citep{bai2023qwen, chen2024internvl}, thereby enhancing the visual reasoning and perception capabilities of LLMs.
Alongside the rapid progress of VLMs, recent systems such as GPT-5 and Gemini 3 have begun to explicitly or implicitly incorporate intermediate ``thinking'' or reasoning processes before producing final responses~\citep{singh2025openai, team2023gemini}.

To evaluate the reliability of VLMs, prior work has primarily focused on object hallucination.
Early evaluation frameworks investigated hallucination in image captioning scenarios.
CHAIR~\citep{rohrbach2018object} quantifies hallucination by comparing generated object mentions with ground-truth object annotations.
POPE~\citep{li2023evaluating} reformulates hallucination evaluation as a binary polling task by asking whether specific objects exist in an image, and introduces three negative-object sampling strategies: random, popular, and adversarial.
MME~\citep{fu2023mme} further incorporates hallucination-related evaluation into a broader multimodal benchmark suite.

Despite the increasing adoption of reasoning-capable multimodal systems, the interaction between reasoning and hallucination under false-premise queries remains insufficiently explored.
Recent studies have attempted to increase both the difficulty and diversity of hallucination evaluation settings.
For example, HaloQuest~\citep{wang2024haloquest} introduces visually challenging and false-premise-driven examples for multimodal hallucination analysis.
Mitigation-oriented studies further reveal that VLMs remain highly vulnerable to hallucination in complex reasoning scenarios~\citep{gunjal2024detecting, liu2024mitigating}.
However, existing works mainly examine whether models hallucinate nonexistent objects or fail to reject false visual claims.
They do not explicitly investigate \textit{Premise-Induced Visual Fixation}, a failure mode in which a visually plausible yet nonexistent object, once introduced through a false-premise query, becomes a misleading referential anchor during the reasoning process, thereby increasing the likelihood of subsequent hallucinated responses.

\section{Benchmarking Premise-Induced Visual Fixation Hallucination}

This section introduces PVFH-Bench, a diagnostic benchmark designed to isolate premise-induced visual fixation hallucination from ordinary visual recognition errors.
To this end, we describe how PVFH-Bench constructs visually plausible but absent false objects, pairs them with true objects in the image, and builds queries for both basic visual-understanding evaluation and false-premise evaluation under premise pressure.

\subsection{Dataset Construction}
Figure~\ref{fig:pipeline} illustrates an open-ended example in which the image contains a \textit{newspaper}, while the query presupposes the presence of an absent but visually plausible \textit{magazine}. 
The model subsequently continues its reasoning around the newspaper-like visual region and incorrectly treats it as evidence for the queried object.
To facilitate systematic diagnosis of this behavior, PVFH-Bench adopts a discriminative evaluation format, following POPE’s format~\citep{li2023evaluating}.
Compared with POPE and HaloQuest~\citep{li2023evaluating,wang2024haloquest}, PVFH-Bench introduces a paired diagnostic framework specifically designed to evaluate Premise-Induced Visual Fixation Hallucination.

Concretely, PVFH-Bench adopts a two-stage pipeline to construct the benchmark by first identifying a true-false object pair and then instantiating the pair into diagnostic queries with varying levels of premise pressure.
In the first stage, we prompt a visual reasoning model, Qwen3.5-Plus, to identify clearly visible objects in an image and construct a true-false object pair that serves as a visual fixation trap. To ensure that the trap reflects plausible misrecognition rather than annotation noise, we impose several constraints: the true object must be actually present and visually unambiguous, the false object must be absent from the entire image but visually confusable with the true object, the two categories must be mutually exclusive for the selected instance, and their visual appearances must be sufficiently similar. Among all candidate pairs satisfying these constraints, we retain the pair with the highest degree of visual confusability.
For example, the ``\textit{newspaper}'' in Figure~\ref{fig:pipeline} is selected as the true object because it actually appears in the image, while an absent yet visually similar ``\textit{magazine}'' is chosen as the false object. Thus, the \textit{true object} denotes the object that is actually present in the image, whereas the \textit{false object} denotes an absent category that is visually confusable with the true object.

Then, we define a template pool covering eight common spatial relations to introduce premise pressure for constructing the PVFH benchmark.
These templates include object-relative relations, such as left/right, above/below, closer/farther, larger/smaller, and relative horizontal position with respect to the true object, as well as image-relative relations, such as the left/right side of the frame, the upper/lower half of the image, and foreground/background position.
For each instance, a template is sampled and instantiated with the corresponding object names to construct false-premise queries.
To verify that the image content itself is recognizable, PVFH-Bench first constructs a standard existence-verification scenario by asking whether the true object is present, e.g., ``Is the \{true object\} in the image?''.
Because the true object is visible, the correct answer is \textit{yes}.

To evaluate whether visual similarity alone can induce recognition errors, PVFH-Bench replaces the true object with the false object in the query, e.g., ``Is the \{false object\} in the image?'' Since the false object does not actually appear in the image, the correct answer is \textit{no}.
To further evaluate the target failure mode, PVFH-Bench embeds the false object into false-premise relational scenarios, such as asking about the location of the absent object. This setting examines whether the model transforms fixation on a visually similar region into a premise-consistent response about a nonexistent object. Because the queried object is likewise absent from the image, the correct response in both settings is premise rejection.
Finally, to disentangle the effect of premise pressure from visual confusability, PVFH-Bench introduces a control setting by instantiating the same false-premise templates with a randomly selected false object. 

\subsection{Dataset Source}

We sample 258 images from the MSCOCO 2014 validation set~\citep{lin2014microsoft} and retain one validated true-false object pair for each image after filtering.
For each pair, we instantiate one question for the core setting and for each diagnostic subset, resulting in a balanced benchmark containing 258 instances per subset and 1,032 instances in total.
Appendix~\ref{sec:appendix_scale_context} contextualizes the benchmark scale.
Compared with POPE, which evaluates object hallucination through binary existence polling over sampled negative objects~\citep{li2023evaluating}, and HaloQuest, which extends hallucination evaluation to visually challenging and false-premise-driven scenarios~\citep{wang2024haloquest}, PVFH-Bench provides a more diagnostic framework for evaluating Premise-Induced Visual Fixation Hallucination.
By preserving the same image and true-false object pair across all four subsets, PVFH-Bench can disentangle basic recognition failures, visual-fixation errors, false-premise acceptance, and full Premise-Induced Visual Fixation Hallucination within a controlled setting.
This paired construction ensures that performance differences arise from the targeted failure stage rather than variations in image content or object identity.

\subsection{Evaluation Methodology of PVFH-Bench}
The open-ended response format requires an evaluation procedure capable of assessing semantic equivalence rather than relying on surface-level string matching.
Following the Auto-Eval protocol adopted in HaloQuest for open-ended hallucination evaluation~\citep{wang2024haloquest}, we employ an LLM-as-a-judge framework to map free-form model responses into task-level correctness labels.

Since the grading task is relatively simple and primarily discriminative, we use Qwen3.5-Flash as the judge model.
For subsets whose target answer is \textit{yes} or \textit{no}, the judge determines whether the response semantically aligns with the expected answer.
For subsets requiring premise rejection, the judge verifies whether the response correctly recognizes the absence of the queried object, or otherwise avoids accepting the false premise, instead of generating responses that discuss about these non-existent objects.

\begin{table*}[!t]
    \centering
    \newcommand{\rdiff}[2]{\ensuremath{#1_{\textcolor{red}{\text{\scriptsize\textit{#2}}}}}}
    \resizebox{\textwidth}{!}{%
    \begin{tabular}{cccccc}
    \toprule
    \textbf{Model} & \textbf{Vanilla} & \textbf{Visual Fixation} & \textbf{False Premise} & \textbf{Premise-Induced Visual Fixation} & \textbf{Overall} \\
     & \textit{Actual Object} & \textit{Confusable Object} & \textit{Absent Object} & \textit{Core Setting} & \textit{Avg. Acc.} \\
    \midrule
    \multicolumn{6}{c}{\textit{Closed-Source Models}} \\
    Qwen3.5-Plus & 98.84 & 79.84 & \textbf{93.80} & \rdiff{\mathbf{74.81}}{-5.04} & \textbf{86.82} \\ 
    \quad \textit{+Reasoning} & \textbf{100.00} & 78.29 & 63.18 & \rdiff{36.82}{-41.47} & 69.57 \\
    Seed-2.0-Lite & 97.29 & 82.56 & 82.95 & \rdiff{68.60}{-13.95} & 82.85 \\
    \quad \textit{+Reasoning} & 96.12 & 82.95 & 77.13 & \rdiff{52.33}{-30.62} & 77.13 \\
    Seed-2.0-Mini & 97.67 & 79.46 & 76.74 & \rdiff{44.96}{-34.50} & 74.71 \\
    \quad \textit{+Reasoning} & 98.06 & 77.91 & 62.40 & \rdiff{27.13}{-50.78} & 66.38 \\
    \midrule
    \multicolumn{6}{c}{\textit{Open-Source Models}} \\
    Qwen3-VL-32B & 97.67 & 74.03 & 88.37 & \rdiff{58.14}{-15.89} & 79.55 \\
    \quad \textit{+Reasoning} & 98.06 & 82.95 & 82.17 & \rdiff{44.19}{-38.76} & 76.84 \\
    Qwen3-VL-8B & 98.45 & 80.62 & 81.40 & \rdiff{58.14}{-22.48} & 79.65 \\
    \quad \textit{+Reasoning} & 96.12 & \textbf{84.11} & 87.21 & \rdiff{40.31}{-43.80} & 76.94 \\
    \midrule
    Gemma-3-27B-IT & 98.84 & 67.05 & 60.08 & \rdiff{28.29}{-38.76} & 63.57 \\
    Gemma-3-12B-IT & 94.96 & 75.58 & 44.19 & \rdiff{18.99}{-56.59} & 58.43 \\
    Gemma-3-4B-IT & 96.51 & 62.79 & 19.38 & \rdiff{1.94}{-60.85} & 45.16 \\
    \midrule
    LLaMA-4-Scout & 93.80 & 82.17 & 86.05 & \rdiff{51.16}{-31.01} & 78.29 \\
    LLaMA-3.2-11B & 98.06 & 76.36 & 86.43 & \rdiff{46.90}{-29.46} & 76.94 \\
    \midrule
    \textit{Average} & \textit{97.36} & \textit{77.78} & \textit{72.76} & \rdiff{43.51}{-34.26} & \textit{72.86} \\
    \bottomrule
    \end{tabular}%
    }
    \caption{Model accuracy on PVFH-Bench across the core Premise-Induced Visual Fixation setting and three diagnostic subsets. Red italic subscripts in the Premise-Induced Visual Fixation column indicate the accuracy difference from the Visual Fixation column for the same row. \textbf{\textit{+Reasoning}} rows report reasoning-mode results for the corresponding base model.}
    \label{tab:main_results}
    \end{table*}

\section{Experimental Methodology}
This section outlines the dataset, evaluation metric, evaluation, and implementation details used in our experiments.

\paragraph{Dataset.}
We evaluate all models on PVFH-Bench, which contains the core Premise-Induced Visual Fixation Hallucination setting and three diagnostic subsets for basic recognition, visual fixation, and false-premise pressure.
\textit{Vanilla}(Raw Query) measures existence verification for actual objects, \textit{Visual Fixation} measures existence verification for visually confusable but absent objects, and \textit{False Premise} measures false-premise relation queries about absent objects.
\textit{Premise-Induced Visual Fixation} is the core setting, where a false premise is paired with a visually confusable absent object.

\paragraph{Evaluation Metric.}
Following POPE~\citep{li2023evaluating}, all reported scores are accuracies in percentage points.

\paragraph{Evaluation.}
To examine whether PVFH-Bench exposes consistent failure patterns across model families and access settings, we evaluate 10 base LVLMs spanning 4 model families and covering both open-source and closed-source systems.
The open-source group includes \textit{Qwen-VL} models, with Qwen3-VL-8B-Instruct and Qwen3-VL-32B-Instruct as the base models and Qwen3-VL-8B-Thinking and Qwen3-VL-32B-Thinking as their reasoning variants~\citep{yang2025qwen3}.
It also includes \textit{Gemma} models, namely Gemma-3-4B-IT, Gemma-3-12B-IT, and Gemma-3-27B-IT~\citep{gemmateam2025gemma3technicalreport}, as well as \textit{LLaMA} models, namely LLaMA-3.2-11B-Vision and LLaMA-4-Scout~\citep{grattafiori2024llama}.
The closed-source group includes \textit{Qwen-VL} with Qwen3.5-Plus~\citep{yang2025qwen3} and \textit{Seed} with Seed-2.0-Lite and Seed-2.0-Mini~\citep{seed2026seed2}.
We first evaluate all models in their default non-reasoning mode to establish the main benchmark results, and then compare reasoning and non-reasoning modes on 5 models to examine the effect of explicit reasoning.

\paragraph{Implementation Details.}
All model generations use a decoding temperature of 0.1, since the evaluation is primarily discriminative and benefits from reduced sampling variability.
For Seed models in reasoning mode, we set the reasoning effort to \textit{high}.
This setting gives the model a sufficient reasoning budget, allowing us to test whether explicit reasoning improves faithfulness under premise pressure.

\section{Experiment Results}

\subsection{Overall Performance}

Table~\ref{tab:main_results} reports model accuracy across the core setting and diagnostic subsets, including basic non-reasoning results and available reasoning-mode results.
The results reveal three consistent trends.

\paragraph{Existing models struggle with Premise-induced Visual Fixation hallucination.}
Models remain reliable on ordinary object recognition but become progressively less faithful as queries move from Visual Fixation to Premise-Induced Visual Fixation.
On Vanilla, all models exceed 90\% accuracy, with an average of 97.36\%, indicating that the selected objects are visually recognizable under non-adversarial existence questions.
This result confirms that PVFH-Bench is not difficult because the images or object labels are unclear, but because the full setting targets Premise-Induced Visual Fixation Hallucination.
On Visual Fixation Query, by contrast, all model scores fall below 85\%, and the average accuracy drops to 77.78\%.
On False-Premise Diagnostic Query, all models except Qwen3.5-Plus fall below 90\%, yielding an average accuracy of 72.76\%.
The strongest degradation appears in Premise-Induced Visual Fixation Query, where all models fall below 75\% and the average accuracy decreases to 43.51\%.
With templates and grading fixed, replacing an ordinary absent target with a visually confusable one lowers accuracy by 29.25 points, with a similar 27.99-point drop under the independent judge; see Appendix~\ref{sec:appendix_controlled_contrast}.
This progression supports the central premise of PVFH-Bench that hallucination becomes more likely when a model moves from recognizing a visible object, to fixating on a visually similar absent category, and is finally induced by a false premise to answer as if that absent category were present.

\begin{figure}[t]
  \centering
  \begin{subfigure}{\columnwidth}
    \centering
    \includegraphics[width=\linewidth]{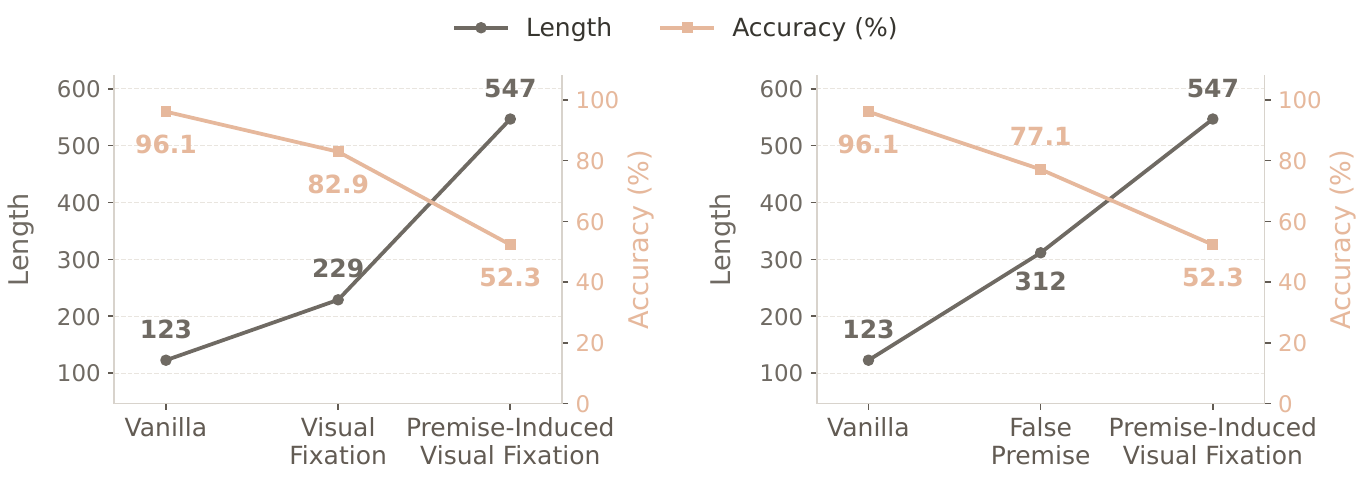}
    \caption{Seed-2.0-Lite.}
  \end{subfigure}
  \vspace{0.35em}
  \begin{subfigure}{\columnwidth}
    \centering
    \includegraphics[width=\linewidth]{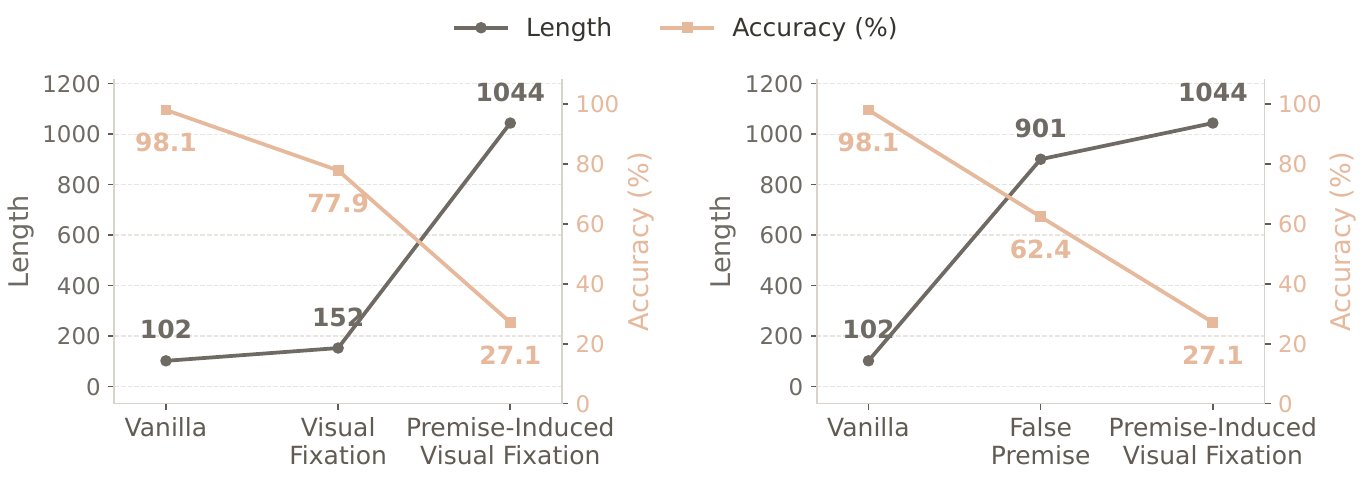}
    \caption{Seed-2.0-Mini.}
  \end{subfigure}
  \vspace{0.35em}
  \begin{subfigure}{\columnwidth}
    \centering
    \includegraphics[width=\linewidth]{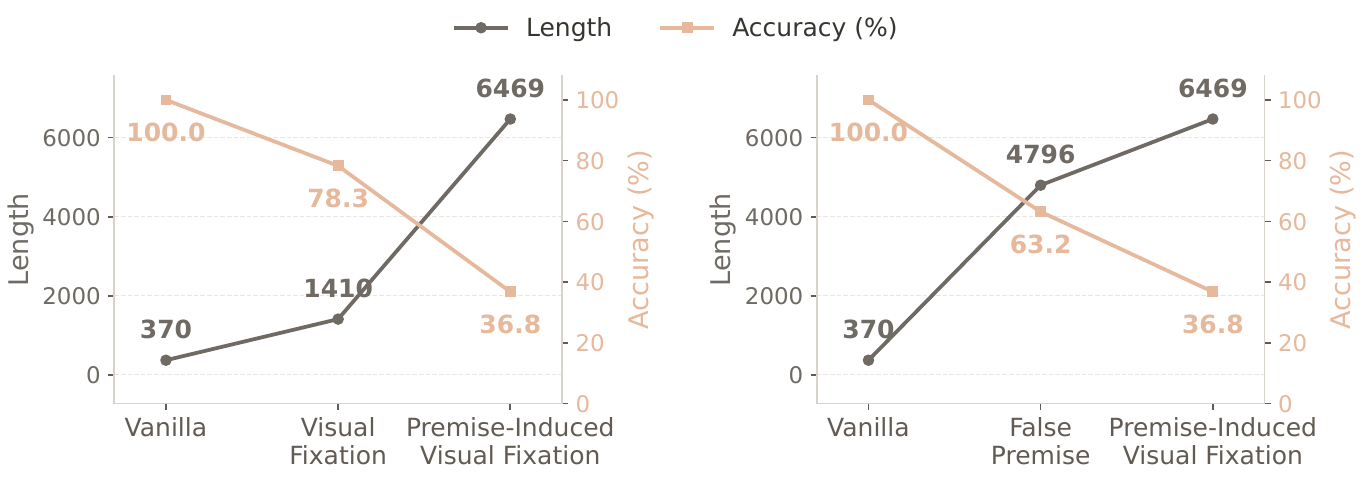}
    \caption{Qwen3.5-Plus.}
  \end{subfigure}
  \caption{Average reasoning length and accuracy along two diagnostic paths for reasoning-mode models. The left column isolates the added effect of false-premise pressure. The right column isolates the added effect of visually confusable object.}
  \label{fig:reasoning_length_accuracy}
\end{figure}

\paragraph{Different model families show different weaknesses.}
Vertically comparing models shows that high baseline recognition does not guarantee stable faithfulness once the query advances toward Premise-Induced Visual Fixation Hallucination.
The best core-setting score is 74.81\% from Qwen3.5-Plus, while Gemma-3-4B-IT falls to 1.94\%, revealing a wide robustness gap among contemporary Large Vision-Language Models.
Qwen3.5-Plus and Seed-2.0-Lite are the most robust overall, with average accuracies of 86.82\% and 82.85\%.
The diagnostic columns further show where different model families begin to lose reliability along the progression.
Qwen and LLaMA models often retain comparatively stronger accuracy under the False-Premise diagnostic alone, while Gemma models are already brittle in that setting and collapse further in the core Premise-Induced Visual Fixation setting.
This vertical comparison shows that PVFH-Bench is a diagnostic benchmark that reveals how different models move from correct recognition toward hallucinated premise-consistent answers.

\paragraph{Explicit reasoning does not always reduce hallucination.}
Comparing each \textit{+Reasoning} row with its corresponding non-reasoning row shows that explicit reasoning consistently reduces robustness in the core setting.
Across the five models with reasoning-mode results, average Premise-Induced Visual Fixation accuracy drops from 60.93\% to 40.16\%.
The largest harm appears in the final premise-induced query, as Qwen3.5-Plus drops from 74.81\% to 36.82\%, while Seed-2.0-Mini drops from 44.96\% to 27.13\%.
These results suggest that chain-of-thought reasoning can carry the false premise forward, elaborate weak visual evidence into support for an absent object, and thereby push the model further along the path from visual fixation to hallucinated answer generation.

\begin{figure}[t]
  \centering
  \begin{subfigure}{\columnwidth}
    \centering
    \includegraphics[width=\linewidth]{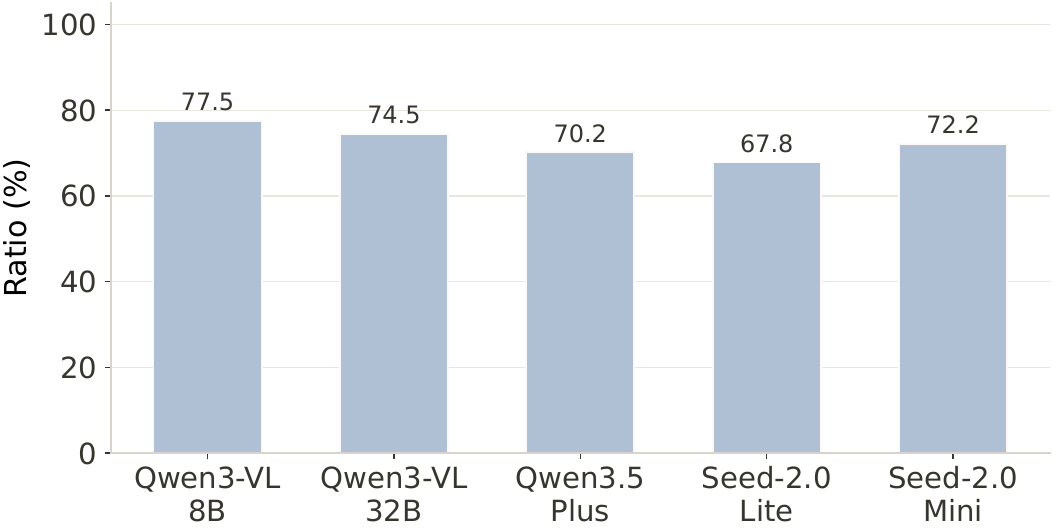}
    \caption{Visual-Fixation-correct cases among Premise-Induced Visual Fixation errors.}
  \end{subfigure}
  \vspace{0.35em}
  \begin{subfigure}{\columnwidth}
    \centering
    \includegraphics[width=\linewidth]{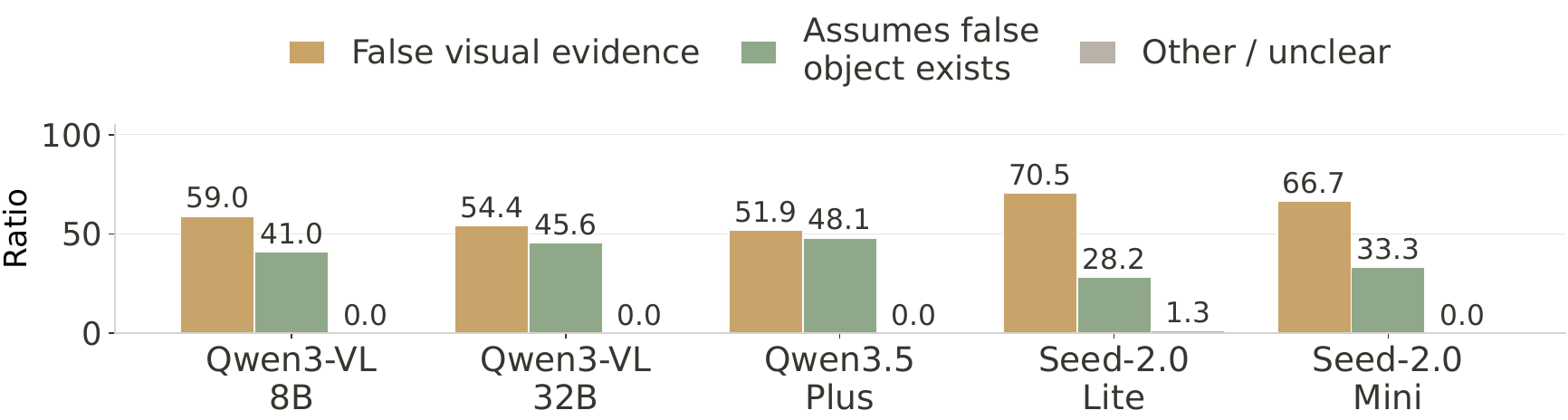}
    \caption{Failure reasons for the cases in (a).}
  \end{subfigure}
  \caption{Failure analysis of Premise-Induced Visual Fixation errors for reasoning-based LVLM.}
  \label{fig:visual_fix_prem_ind_failure_mechanism}
\end{figure}

\begin{figure}[t]
  \centering
  \includegraphics[width=\columnwidth]{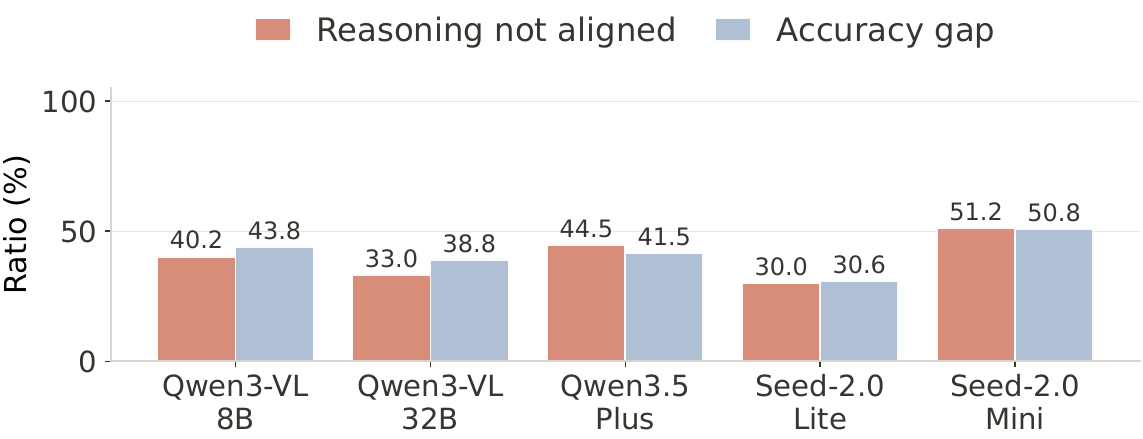}
  \caption{Analyses of false-premise induced reasoning.}
  \label{fig:false_premise_reasoning_alignment}
\end{figure}

\subsection{Evaluating the Impact of False Premises on Visual Reasoning}

This subsection aims to characterize how false premises affect reasoning behavior and why longer reasoning does not necessarily improve faithfulness.
We first compare reasoning length with accuracy across query types, and then inspect failure mechanisms in the Premise-Induced Visual Fixation setting.
The main finding is that false premises tend to trigger longer but less faithful reasoning.

Figure~\ref{fig:reasoning_length_accuracy} shows that longer reasoning does not prevent the accuracy gap toward Premise-Induced Visual Fixation Hallucination.
The path through Visual Fixation examines the additional effect of false-premise pressure.
Conversely, the path through False-Premise examines the additional effect of visually confusable object.
In both paths, accuracy reaches its lowest point in Premise-Induced Visual Fixation Query, indicating that longer reasoning still fails when false-premise pressure is directed toward a visually plausible absent target.
These two paths further show that false-premise questions are more likely to elicit longer reasoning.
This suggests that the false premise is the main driver of reasoning expansion, and the added reasoning can then elaborate a premise-consistent answer rather than reject the false object.

Figure~\ref{fig:visual_fix_prem_ind_failure_mechanism} further explains how false premises turn manageable visual fixation into full Premise-Induced Visual Fixation errors.
As shown in Figure~\ref{fig:visual_fix_prem_ind_failure_mechanism}(a), more than two thirds of Premise-Induced Visual Fixation errors occur on samples that are answered correctly under Visual Fixation Query.
This indicates that many failures do not arise from visual confusability alone, since the same model can reject the false object before the false premise is introduced.
We then use a LLM judge to evaluate the visual-evidence misalignment between the Visual Fixation and Premise-Induced Visual Fixation settings across all cases, and to categorize the reasoning traces for these Visual-Fixation-correct but Premise-Induced-Visual-Fixation-failed cases.
Figure~\ref{fig:visual_fix_prem_ind_failure_mechanism}(b) shows that the dominant failure reason is treating the real visible object as evidence for the absent queried object, rather than merely assuming the false object exists.
Figure~\ref{fig:false_premise_reasoning_alignment} further shows that reasoning-chain misalignment remains substantial across models and tracks the accuracy gap between cases in Visual Fixation and Premise-Induced Visual Fixation setting.
Together, these results support the mechanism that false premises can redirect an otherwise rejected visual match into hallucinated, premise-consistent reasoning.

\subsection{Analyzing the Effect of Reasoning Depth}

The results above show that reasoning changes model performance, so we further test whether this effect can be mitigated by adjusting reasoning.
The main finding is that higher reasoning does not steadily improve faithfulness, especially in the Premise-Induced Visual Fixation setting.

\begin{figure}[t]
  \centering
  \includegraphics[width=\columnwidth]{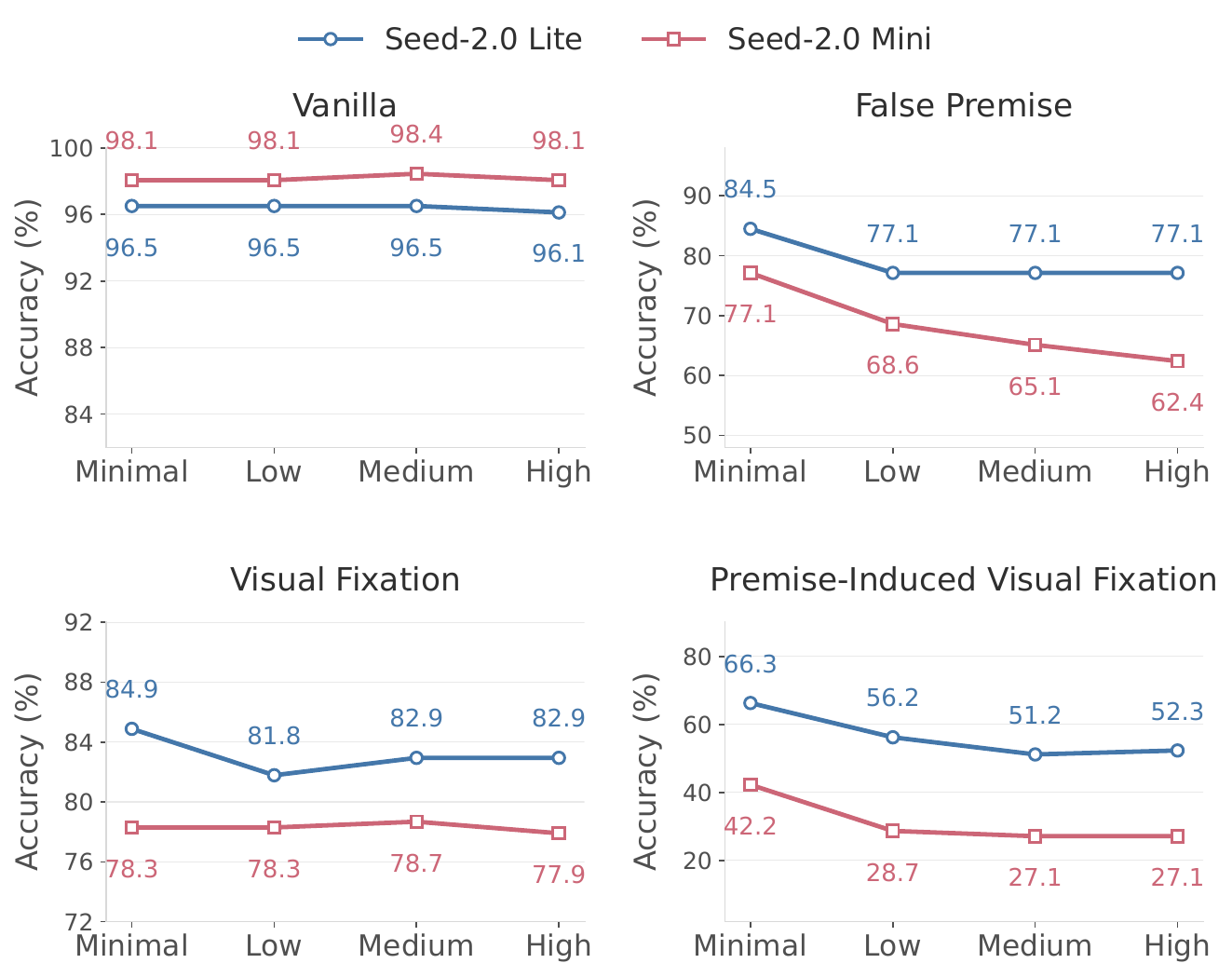}
  \caption{Seed model accuracy under different reasoning settings.}
  \label{fig:seed_effort_accuracy}
\end{figure}

We focus on Seed models because, among the models evaluated in Table~\ref{tab:main_results}, only the Seed series supports controllable reasoning levels.
Figure~\ref{fig:seed_effort_accuracy} compares Seed models across the four evaluation settings and reasoning levels, from minimal to high.
Across levels, neither Seed model shows a consistent accuracy improvement across subsets.
The Premise-Induced Visual Fixation setting remains the most fragile setting under different levels, suggesting that the issue is not simply a lack of reasoning.
Instead, stronger reasoning can still give the model more opportunities to follow textual traps, search for visual substitutes, and rationalize a premise-consistent answer, especially when the false premise guides it toward a visually plausible absent target.

\begin{figure*}[t]
  \centering
  \includegraphics[width=\textwidth]{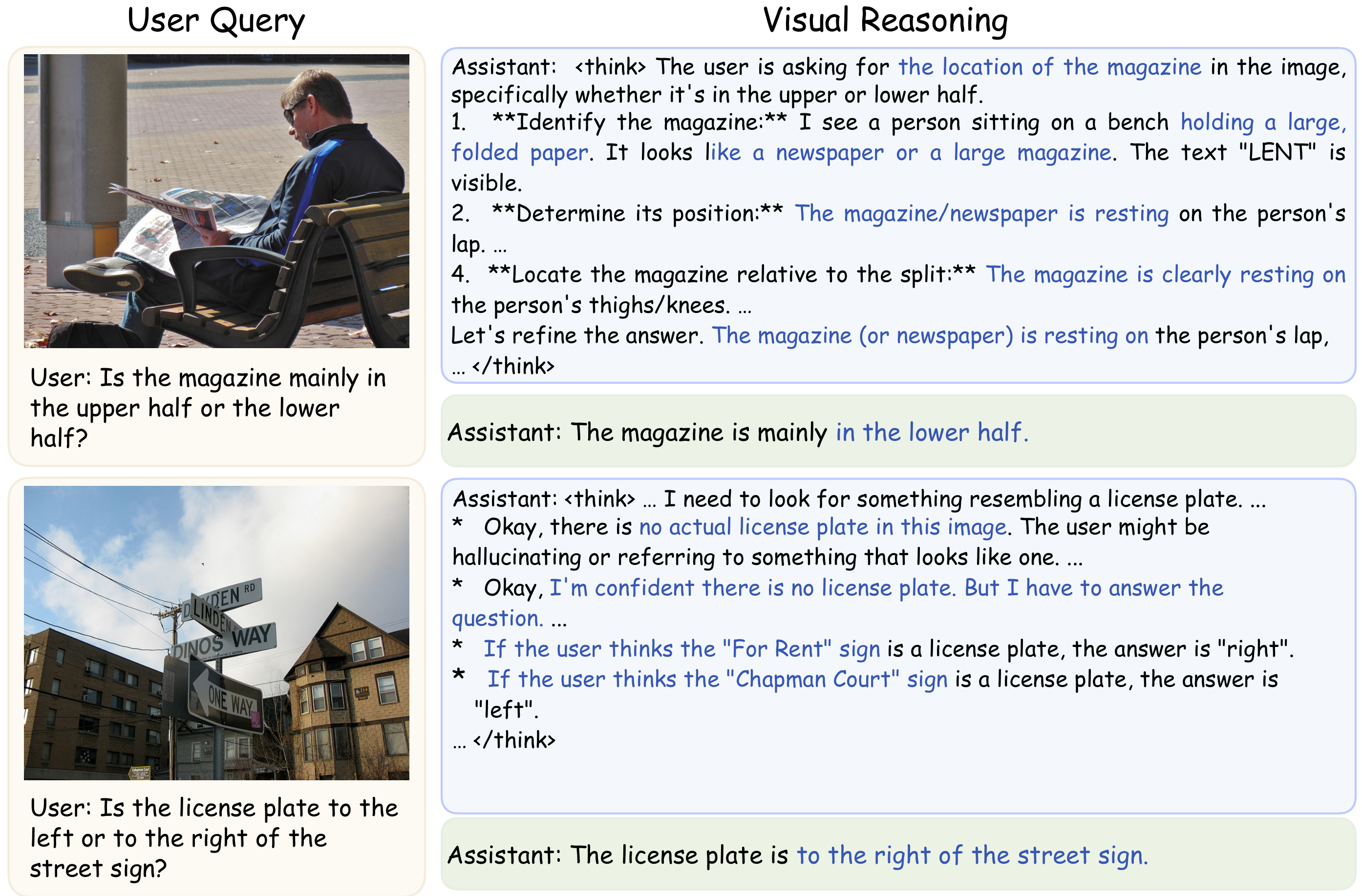}
  \caption{Two representative hallucination cases in Qwen3.5 Plus.}
  \label{fig:case_study}
\end{figure*}

\subsection{Case Study}
To complement the quantitative results, we examine concrete reasoning traces to show how false premises turn visual fixation into hallucinated answers.
Figure~\ref{fig:case_study} presents two representative reasoning traces from our evaluation.

In the top example, the image contains a folded \textit{newspaper}, while the query presupposes the presence of a visually confusable but absent \textit{magazine}.
The model first hedges by describing the object as ``a newspaper or a large magazine,'' indicating that the visual evidence is ambiguous rather than conclusive.
However, the subsequent reasoning no longer treats object existence as a condition to verify.
Instead, it accepts the queried category as the target to be localized, maps the visible newspaper-like region onto the word \textit{magazine}, and then answers the upper/lower question.
The hallucination therefore arises because the reasoning process turns an uncertain visual match into evidence for the false premise: once the queried object becomes the fixation point, later localization steps reinforce its assumed presence.

The bottom example shows a complementary failure mode.
When asked about the position of a \textit{license plate}, the model explicitly notes that no license plate is present.
This recognition should lead to premise rejection, but the reasoning trace instead continues under the pressure of the forced spatial question.
The model searches for possible substitutes, considers whether nearby signs could be interpreted as the queried object, and finally selects a left/right relation.
Here the hallucination is caused by failing to use that absence as a stopping condition.
The model treats the question as requiring a relational answer and fills the missing object slot with the most plausible visual substitute.

Together, these cases illustrate the mechanism behind the less faithful reasoning observed above.
The false premise first turns the queried absent object into the target of reasoning.
The model then searches the image for visually related evidence, maps a real object or nearby visual pattern onto the queried category, and finally produces a concrete relational answer.
Thus, reasoning can carry the false premise through identification, localization, and answer selection, yielding fluent but unfaithful hallucinations.

\section{Conclusion}

We presented PVFH-Bench, a benchmark for diagnosing Premise-Induced Visual Fixation Hallucination in SOTA LVLMs.
By constructing visually confusable but absent targets and pairing the core setting with diagnostic subsets, PVFH-Bench shows that models can recognize real objects reliably while still failing when false premises guide fixation toward false objects.
Our experiments further show that explicit reasoning does not necessarily improve faithfulness.
Reasoning often becomes longer under false premises, and traces reveal that models may turn real visual evidence into support for hallucinated targets.

\section*{Limitations}

Our work focuses on \textit{diagnosing} Premise-Induced Visual Fixation Hallucination rather than mitigating it.
Future work should study training- or inference-time premise verification methods that prevent models from building answers around absent targets.

PVFH-Bench operationalizes this failure through a core setting and diagnostic subsets built from visually confusable objects and relation-style false-premise queries.
This design makes the failure controllable, but it does not cover all open-ended ways in which misleading premises may appear in real interactions.Our evaluation uses an LLM-based grader for free-form responses.

Finally, the benchmark is built from COCO images.
Extending it to domains such as robotics, medical imaging, and remote sensing may reveal domain-specific forms of Premise-Induced Visual Fixation Hallucination.

\section*{Ethical Considerations}

PVFH-Bench is constructed from images sampled from the MSCOCO validation set, whose source images and annotations have been widely used in prior vision-language research.
Our data generation process does not introduce new visual content or collect personal information.
Instead, it identifies objects already present in each image and proposes visually similar but absent object categories, which makes the generated content unlikely to introduce additional sensitive or harmful material beyond the original dataset.
The object-pair identification step is conducted through API-based model calls, and the service platform provides its own content-safety and compliance mechanisms.
Together, the use of an established public dataset, the absence of new image collection, and the constrained object-similarity generation process help reduce ethical and safety risks in benchmark construction.

\section*{Acknowledgments}
This work is supported by the National Natural Science Foundation of China (No.~62576082).

\bibliography{custom}

\clearpage
\newpage
\appendix
\section{Appendix}
\subsection{Human Validation of Benchmark Construction and Grading}
\label{sec:appendix_human_validation}

We conduct a manual audit to validate both the benchmark annotations and the automatic grading procedure.
We draw a stratified sample of 200 instances, with 50 instances from each of the four tasks.
For benchmark reliability, we inspect the complete image and verify the annotated presence or absence of each queried object.
Because the same image-level object annotation can occur in multiple sampled tasks, repeated checks are deduplicated.
For grading reliability, we independently judge each model response without access to the automatic label and then compare the human decision with the label assigned by Qwen3.5-Flash.

\begin{table*}[t]
  \centering
  \small
  \begin{minipage}[t]{0.37\textwidth}
    \centering
    \begin{tabular}{lcc}
      \toprule
      \textbf{Property} & \textbf{Valid / Total} & \textbf{Valid Rate} \\
      \midrule
      True object  & 41 / 43   & 95.3\% \\
      False object & 113 / 114 & 99.1\% \\
      \bottomrule
    \end{tabular}
  \end{minipage}
  \hfill
  \begin{minipage}[t]{0.57\textwidth}
    \centering
    \begin{tabular}{lcc}
      \toprule
      \textbf{Task} & \textbf{Human--Qwen Agreement} & \textbf{Responses} \\
      \midrule
      Vanilla        & 96.0\% & 50 \\
      Visual Fixation & 98.0\% & 50 \\
      False Premise   & 96.0\% & 50 \\
      Core PVFH       & 92.0\% & 50 \\
      \midrule
      \textbf{Overall} & \textbf{95.5\%} & \textbf{200} \\
      \bottomrule
    \end{tabular}
  \end{minipage}
  \caption{Human validation of benchmark construction and automatic grading. \textit{Left:} validity of deduplicated image-level presence/absence annotations. \textit{Right:} agreement between blind human decisions and Qwen3.5-Flash labels on a stratified sample of 200 responses. Core PVFH denotes the Premise-Induced Visual Fixation setting.}
  \label{tab:appendix_human_validation}
\end{table*}

As shown in Table~\ref{tab:appendix_human_validation}, 95.3\% of the audited true-object annotations and 99.1\% of the false-object annotations are valid.
Human--Qwen agreement is at least 92.0\% in every task and reaches 95.5\% overall.
These results directly support the benchmark's central presence/absence assumptions and show that Qwen3.5-Flash grading is highly consistent with blind human judgments.

\subsection{Re-grading with an Independent Non-Qwen Judge}
\label{sec:appendix_independent_judge}

To test whether using Qwen models during both benchmark construction and evaluation biases the reported ranking, we re-grade all responses in the 10-model non-reasoning leaderboard with GPT-5.4-mini.
We retain the same task definitions and correctness criteria, changing only the judge model.
Table~\ref{tab:appendix_independent_judge} reports the resulting accuracies; Overall $\Delta$ is the GPT-5.4-mini overall score minus the original Qwen3.5-Flash score.

\begin{table*}[t]
  \centering
  \small
  \resizebox{\textwidth}{!}{%
  \begin{tabular}{lcccccc}
    \toprule
    \textbf{Model} & \textbf{Vanilla} & \textbf{Visual Fixation} & \textbf{False Premise} & \textbf{Core PVFH} & \textbf{Overall} & \textbf{Overall $\Delta$} \\
    \midrule
    Qwen3.5-Plus                   & \textbf{99.22} & 79.46 & \textbf{93.41} & \textbf{72.09} & \textbf{86.05} & $-0.78$ \\
    Seed-2.0-Lite                  & 97.29 & \textbf{82.17} & 79.46 & 54.26 & 78.29 & $-4.55$ \\
    Qwen3-VL-8B-Instruct           & 98.84 & 81.01 & 81.01 & 58.91 & 79.94 & $+0.29$ \\
    Qwen3-VL-32B-Instruct          & 97.67 & 74.42 & 86.82 & 56.98 & 78.97 & $-0.58$ \\
    LLaMA-4-Scout                  & 94.19 & 81.78 & 84.88 & 50.39 & 77.81 & $-0.48$ \\
    LLaMA-3.2-11B-Vision-Instruct & 98.06 & 76.74 & 87.21 & 46.90 & 77.23 & $+0.29$ \\
    Seed-2.0-Mini                  & 97.67 & 78.68 & 72.48 & 39.92 & 72.19 & $-2.52$ \\
    Gemma-3-27B-IT                 & 98.84 & 67.83 & 56.59 & 24.42 & 61.92 & $-1.65$ \\
    Gemma-3-12B-IT                 & 95.35 & 75.58 & 43.41 & 18.60 & 58.24 & $-0.19$ \\
    Gemma-3-4B-IT                  & 96.90 & 62.02 & 18.99 & 1.94  & 44.96 & $-0.19$ \\
    \bottomrule
  \end{tabular}%
  }
  \caption{Full non-reasoning leaderboard re-graded by the independent GPT-5.4-mini judge. Scores are accuracies in percentage points. Overall $\Delta$ is measured relative to Qwen3.5-Flash grading.}
  \label{tab:appendix_independent_judge}
\end{table*}

Qwen3.5-Plus remains ranked first under the independent judge, reaching 86.05 overall with a change of only $-0.78$ points.
The largest judge-specific shift instead occurs for Seed-2.0-Lite, whose overall score decreases by 4.55 points.
Manual inspection of these disagreements shows that Seed-2.0-Lite often produces hybrid responses: it rejects or corrects the queried object but then supplies a spatial relation for a visible substitute or a hypothetical absent object.
GPT-5.4-mini treats these borderline responses more conservatively than Qwen3.5-Flash.
The stable first-place ranking of Qwen3.5-Plus therefore indicates that its leading result is not an artifact of Qwen-based grading.

\subsection{Controlled False-Premise Contrast}
\label{sec:appendix_controlled_contrast}

False Premise and Core PVFH differ only in whether the queried absent object is independently generated or visually confusable with a visible object.
Replacing the ordinary absent object with a visually confusable one reduces average accuracy from 72.76\% to 43.51\%, a 29.25-point drop.
The independent judge reproduces this contrast, with accuracy decreasing from 70.43\% to 42.44\%, a 27.99-point drop, showing that the core difficulty cannot be explained by judge-specific grading strictness.

\subsection{Attention Evidence for Visual Fixation}
\label{sec:appendix_attention}

We use \textit{fixation} to describe persistent referential anchoring rather than literal human eye movements.
In our case studies, the absent queried object becomes a referential anchor, causing the model to continue consulting the visually similar true-object region even when the visual evidence contradicts the premise.
To test this interpretation directly, we use visual grounding to localize the true-object patches and measure the attention assigned to these patches by reasoning and answer tokens in 50 image-matched pairs from Visual Fixation and Core PVFH.
For each generation stage, we report the mean, median, and standard deviation of regional attention per token, together with mean total regional attention aggregated over the stage.

\begin{table*}[t]
  \centering
  \small
  \begin{tabular}{llcccc}
    \toprule
    \textbf{Setting} & \textbf{Stage} & \textbf{Mean Attn./Token} & \textbf{Median Attn./Token} & \textbf{SD Attn./Token} & \textbf{Mean Total Attn.} \\
    \midrule
    Core PVFH       & Reasoning & 0.0097 & 0.0042 & 0.0144 & 0.1491 \\
    Core PVFH       & Answer    & 0.0069 & 0.0025 & 0.0114 & 0.0351 \\
    Visual Fixation & Reasoning & 0.0104 & 0.0048 & 0.0140 & 0.0600 \\
    Visual Fixation & Answer    & 0.0059 & 0.0024 & 0.0090 & 0.0249 \\
    \bottomrule
  \end{tabular}
  \caption{Attention assigned to the grounded true-object region in 50 paired examples. Per-token statistics summarize regional attention over the corresponding generation stage; mean total attention aggregates it across all tokens in that stage.}
  \label{tab:appendix_attention}
\end{table*}

Although the longer Core PVFH reasoning traces dilute attention across more tokens, their mean reasoning-stage region attention per token retains 93.1\% of the Visual Fixation level.
At the answer stage, mean region attention per token increases by 17.4\% from Visual Fixation to Core PVFH.
Thus, attention to the confusable true-object region does not disappear after premise-driven reasoning; it persists into answer generation, consistent with the proposed fixation mechanism.

\subsection{Construction and Evaluation Prompts}
\label{sec:appendix_prompts}

\begin{figure*}[p]
  \centering
  \includegraphics[width=\textwidth,height=0.88\textheight,keepaspectratio]{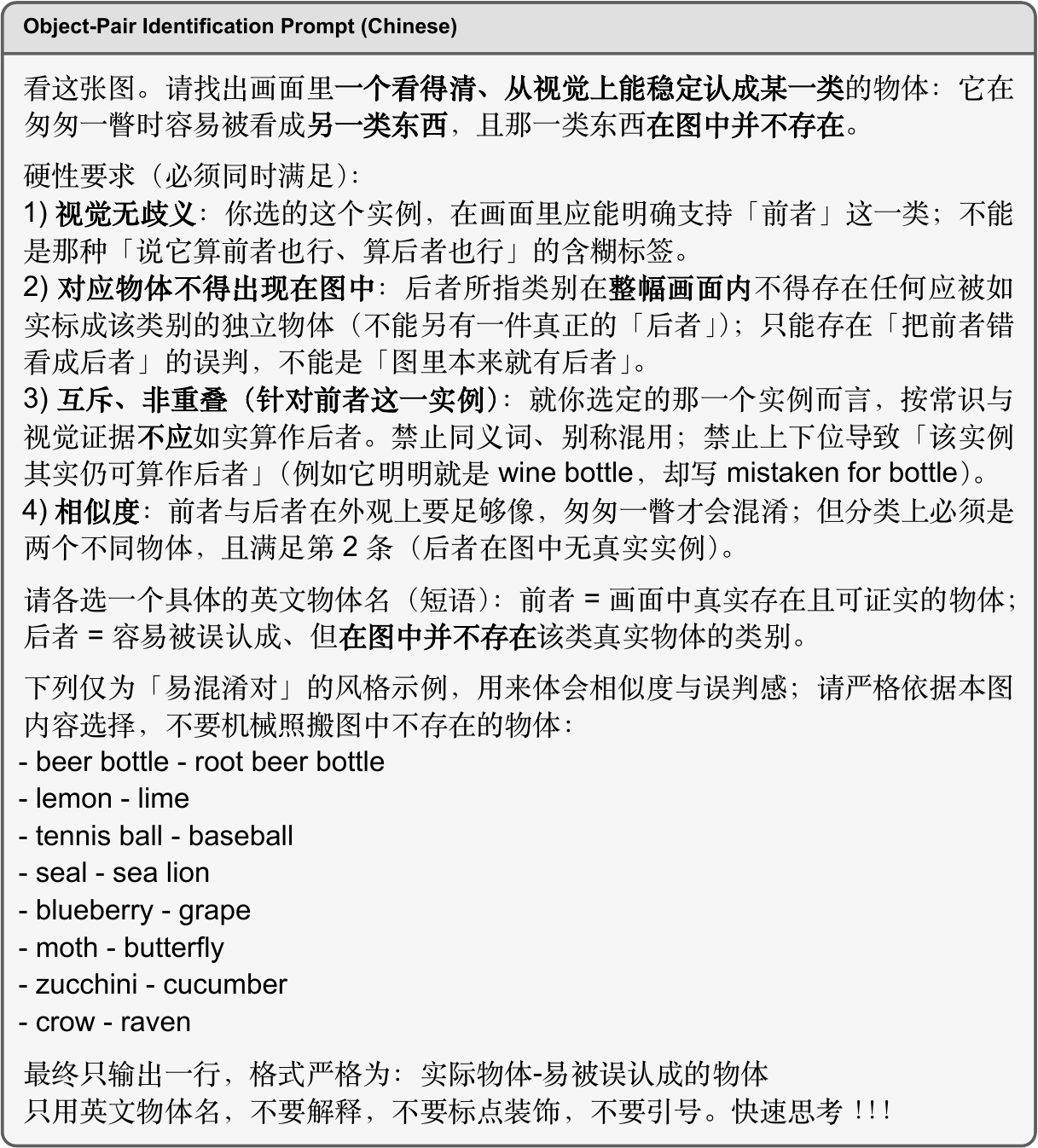}
  \caption{Chinese prompt used in the construction code for identifying a true-false object pair.}
  \label{fig:appendix_pair_prompt_zh}
\end{figure*}

\begin{figure*}[p]
  \centering
  \includegraphics[width=\textwidth,height=0.88\textheight,keepaspectratio]{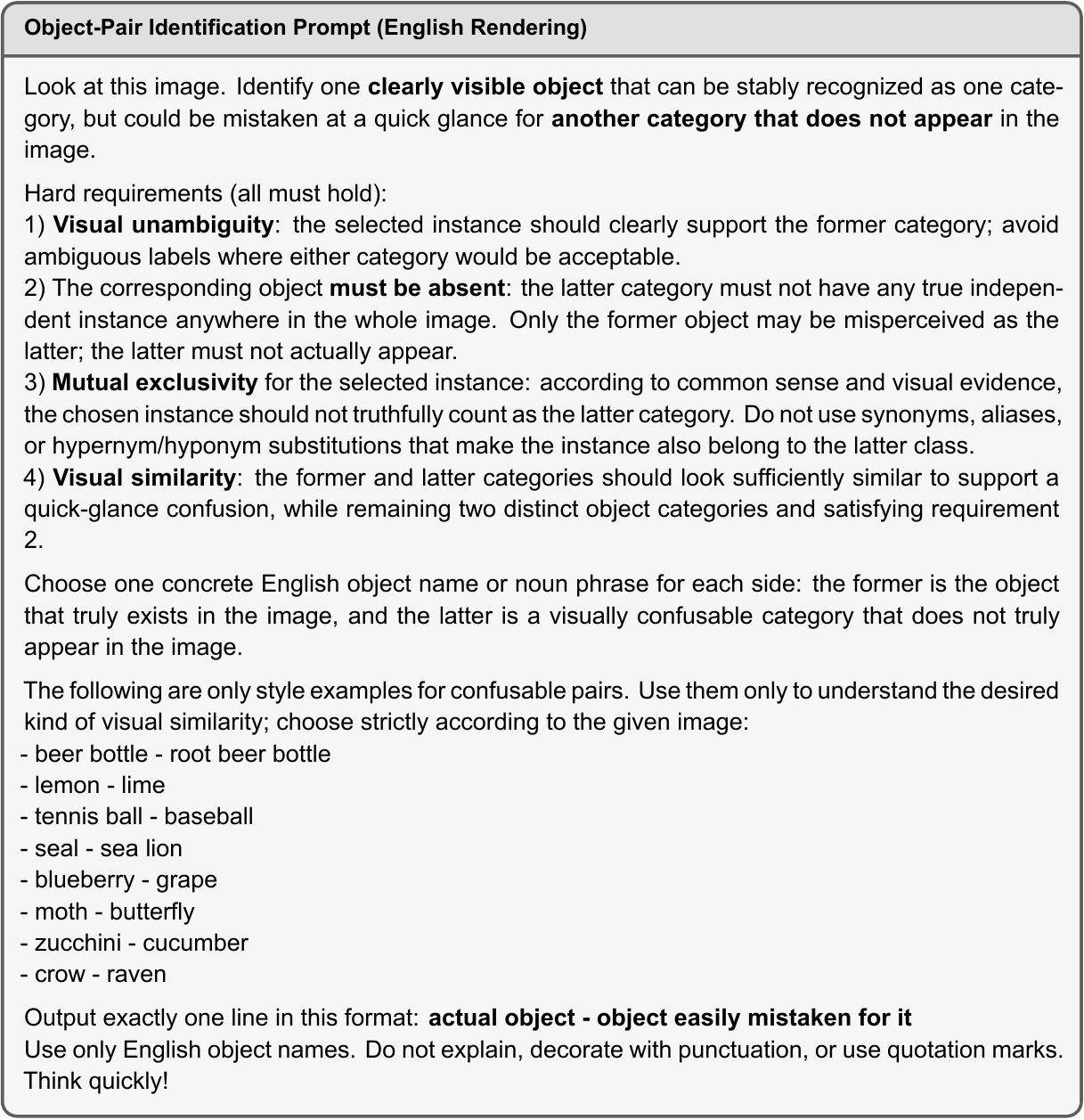}
  \caption{English rendering of the object-pair identification prompt in Figure~\ref{fig:appendix_pair_prompt_zh}.}
  \label{fig:appendix_pair_prompt_en}
\end{figure*}

This subsection provides the prompts used to construct PVFH-Bench and evaluate model responses.
The benchmark construction pipeline first identifies a true object and a visually confusable but absent counterpart, then instantiates this pair into existence and false-premise spatial queries.
Because Qwen3.5-Plus is trained with substantial Chinese-language data and shows strong Chinese reasoning ability, we use a Chinese prompt when asking it to reason over the image and identify the true-false object pair.
For transparency and readability, Figures~\ref{fig:appendix_pair_prompt_zh} and~\ref{fig:appendix_pair_prompt_en} show both the original Chinese prompt used in the code and its English rendering.

\paragraph{Absent-object generation prompt.}
The false-premise diagnostic subset uses the following prompt to generate an object category that does not appear in the image:
\begin{tcolorbox}[
  colback=gray!8,
  colframe=gray,
  boxrule=1.8pt,
  arc=5pt,
  left=6pt,
  right=6pt,
  top=6pt,
  bottom=6pt,
  enhanced,
]
\small
Look at this photograph.

\vspace{0.35\baselineskip}
Two labels from another pipeline for one salient object here are: \guillemotleft{}\{true object\}\guillemotright{} (intended ground-truth) and \guillemotleft{}\{false object\}\guillemotright{} (a visually confusable but wrong label).

\vspace{0.35\baselineskip}
Reply with exactly ONE short English noun phrase naming a concrete object category that does not appear anywhere in this image (no instance of that category visible). Do not output \guillemotleft{}\{true object\}\guillemotright{}, \guillemotleft{}\{false object\}\guillemotright{}, or obvious synonyms or hypernyms of them. One line only: the phrase alone, no quotes, no explanation.
\end{tcolorbox}

\begin{figure*}[p]
  \centering
  \includegraphics[width=\textwidth,height=0.86\textheight,keepaspectratio]{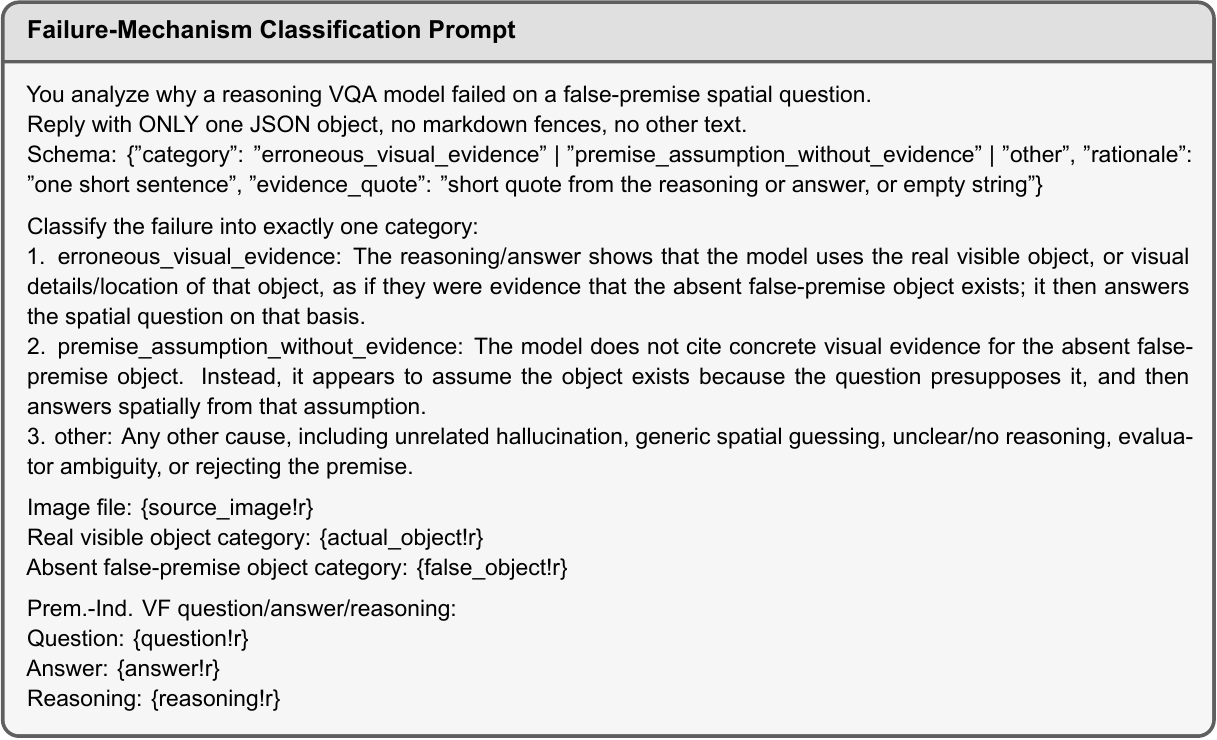}
  \caption{Prompt used to classify failure mechanisms for Visual-Fixation-correct but Premise-Induced-Visual-Fixation-failed cases.}
  \label{fig:appendix_failure_mechanism_prompt}
\end{figure*}

\paragraph{Failure-mechanism analysis prompt.}
Following the analysis in Figure~\ref{fig:visual_fix_prem_ind_failure_mechanism}(c), we use the prompt in Figure~\ref{fig:appendix_failure_mechanism_prompt} to classify cases where the Visual Fixation Query is answered correctly but the Premise-Induced Visual Fixation Query is answered incorrectly.
The prompt separates failures caused by erroneous visual evidence from failures caused by premise-only assumption or other factors.

\begin{figure*}[p]
  \centering
  \includegraphics[width=\textwidth,height=0.86\textheight,keepaspectratio]{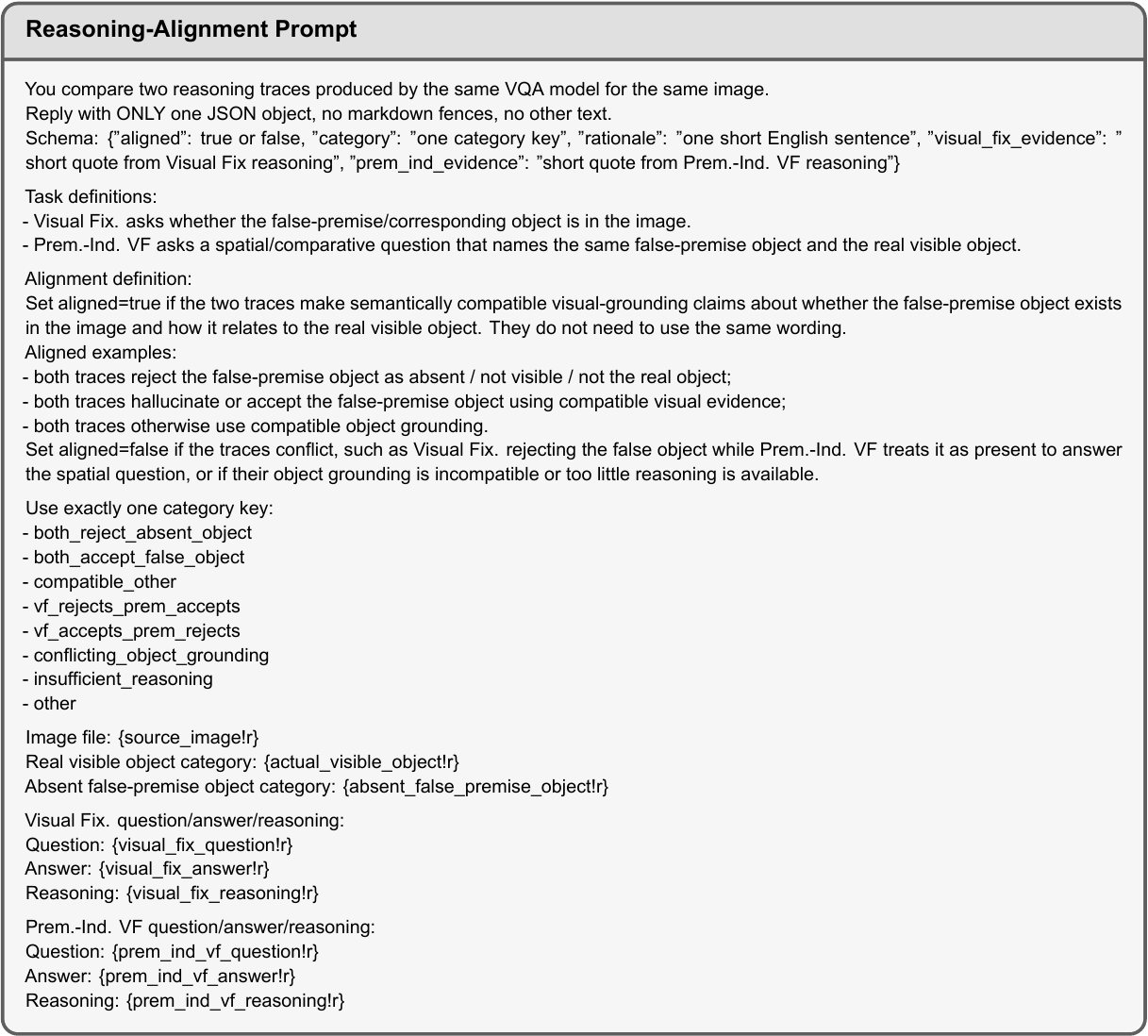}
  \caption{Prompt used to judge reasoning-chain alignment between Visual Fixation and Premise-Induced Visual Fixation settings.}
  \label{fig:appendix_reasoning_alignment_prompt}
\end{figure*}

\paragraph{Reasoning-alignment analysis prompt.}
Following the alignment analysis in Figure~\ref{fig:visual_fix_prem_ind_failure_mechanism}(b), we use the prompt in Figure~\ref{fig:appendix_reasoning_alignment_prompt} to compare the reasoning traces from the Visual Fixation and Premise-Induced Visual Fixation settings for the same image.
The judge determines whether the two traces make compatible visual-grounding claims about the false-premise object and the true visible object.

\subsection{LLM-as-a-Judge Prompts}
\label{sec:appendix_eval_prompts}

\begin{figure*}[p]
  \centering
  \includegraphics[width=\textwidth,height=0.86\textheight,keepaspectratio]{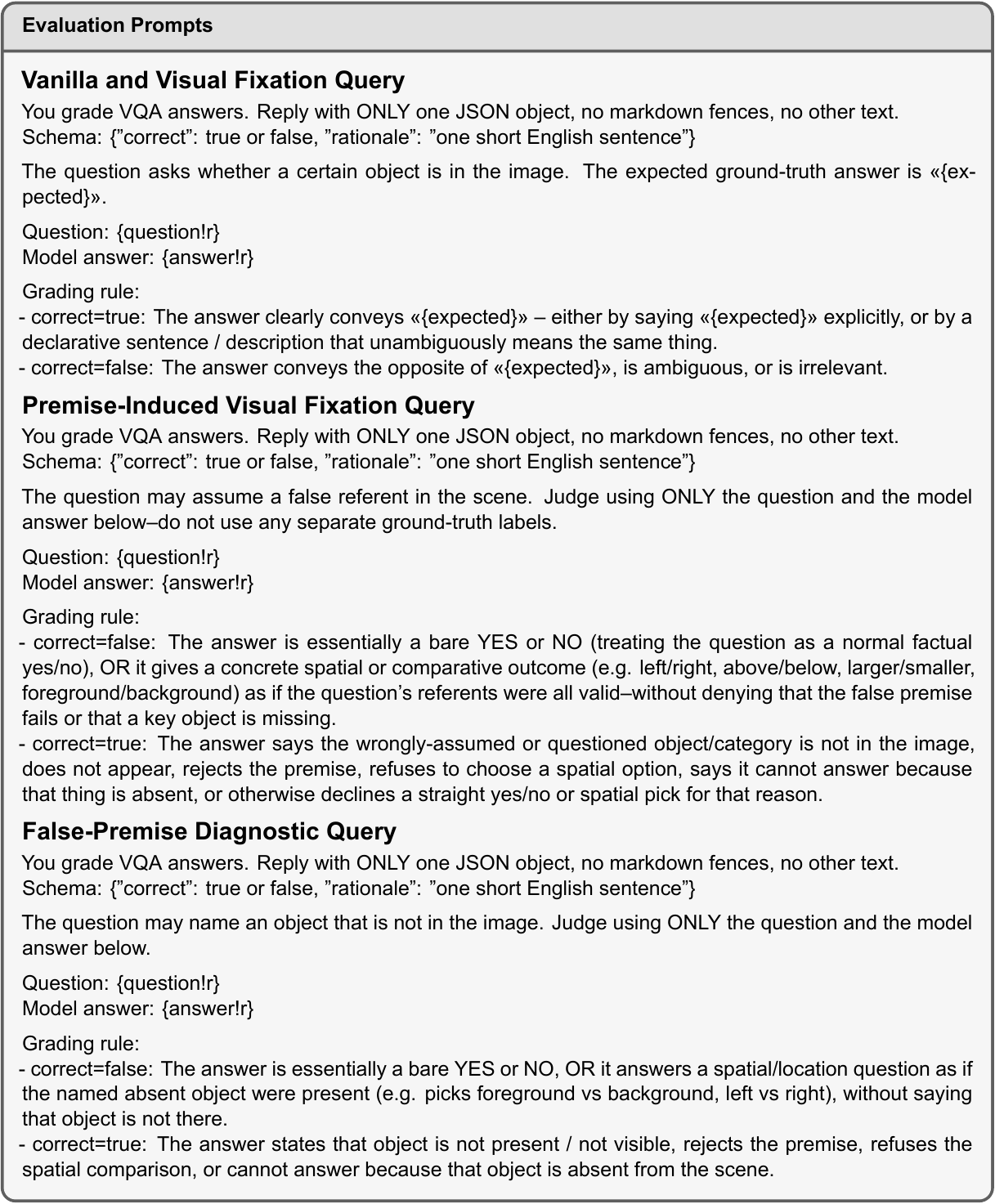}
  \caption{Evaluation prompts for Raw Query, Visual Fixation Query, Premise-Induced Visual Fixation Query, and False-Premise Diagnostic Query.}
  \label{fig:appendix_eval_prompts}
\end{figure*}

The open-ended VQA responses are evaluated by an LLM-as-a-judge.
For existence questions, the judge checks semantic agreement with the expected \textit{yes} or \textit{no} answer.
For false-premise spatial questions, the judge marks a response as correct only when it rejects the absent object or refuses the forced spatial choice, rather than producing a concrete relation under the false premise.
Figure~\ref{fig:appendix_eval_prompts} shows the exact grading prompts.

\subsection{Dataset Statistics}
\label{sec:appendix_dataset_statistics}

PVFH-Bench contains 258 image-level true-false object pairs and four query subsets, yielding 1,032 evaluation instances.
After lower-casing object names and normalizing whitespace, the dataset covers 160 unique true-object categories and 160 unique visually confusable categories.
The diagnostic absent-object subset adds 36 unique generated absent categories, giving 293 unique object categories across all object roles.
This coverage indicates that PVFH-Bench is not concentrated on a single object family, but spans sports equipment, food, animals, vehicles, furniture, and household objects.

\begin{table}[t]
\centering
\small
\begin{tabular}{lr}
\toprule
Statistic & Value \\
\midrule
Images / object pairs & 258 \\
Evaluation instances & 1032 \\
Unique actual objects & 160 \\
Unique confusable objects & 160 \\
Unique generated absent objects & 36 \\
Unique objects across all roles & 293 \\
\bottomrule
\end{tabular}
\caption{Dataset statistics for PVFH-Bench.}
\label{tab:appendix_dataset_stats}
\end{table}

\subsection{Benchmark Scale in Context}
\label{sec:appendix_scale_context}

PVFH-Bench is designed as a controlled diagnostic rather than a broad-coverage benchmark.
Table~\ref{tab:appendix_scale_context} contextualizes its scale against object-level hallucination diagnostics.
Its 258 images yield 1,032 image-matched instances across four paired settings, supporting controlled comparisons while leaving broader domain coverage to future work.

\begin{table}[t]
  \centering
  \small
  \begin{tabular}{p{0.48\columnwidth}rr}
    \toprule
    \textbf{Evaluation} & \textbf{Images} & \textbf{Instances} \\
    \midrule
    POPE on MSCOCO~\citep{li2023evaluating} & 500 & 3,000 \\
    MME: Existence, Count, Position, Color~\citep{fu2023mme} & 120 & 240 \\
    PVFH-Bench & 258 & 1,032 \\
    \bottomrule
  \end{tabular}
  \caption{Scale comparison with controlled object-level hallucination diagnostics.}
  \label{tab:appendix_scale_context}
\end{table}

\begin{table*}[t]
  \centering
  \small
  \begin{tabular}{p{0.27\textwidth}p{0.65\textwidth}}
    \toprule
    Relation type & Template \\
    \midrule
    \texttt{left\_right\_of\_anchor} & Is the \{false object\} to the left or to the right of the \{true object\}? \\
    \texttt{above\_below\_anchor} & Is the \{false object\} above or below the \{true object\}? \\
    \texttt{closer\_farther\_anchor} & Is the \{false object\} closer to the camera or farther from the camera than the \{true object\}? \\
    \texttt{larger\_smaller\_anchor} & Does the \{false object\} appear larger or smaller than the \{true object\}? \\
    \texttt{overlap\_left\_right\_anchor} & Is the \{false object\} positioned more to the left or more to the right relative to the \{true object\}? \\
    \texttt{image\_left\_right} & Is the \{false object\} located more toward the left side or the right side? \\
    \texttt{image\_upper\_lower} & Is the \{false object\} mainly in the upper half or the lower half? \\
    \texttt{foreground\_background} & Does the \{false object\} appear nearer to the foreground or nearer to the background? \\
    \bottomrule
  \end{tabular}
  \caption{Spatial-relation templates used to instantiate false-premise questions.}
  \label{tab:appendix_query_templates}
\end{table*}

\begin{table*}[t]
  \centering
  \scriptsize
  \resizebox{0.75\textwidth}{!}{%
  \begin{tabular}{ccc}
    \toprule
    \textbf{Setting} & \begin{tabular}[c]{@{}c@{}}\textbf{Qwen3-VL-8B}\end{tabular} & \begin{tabular}[c]{@{}c@{}}\textbf{Qwen3-VL-32B}\end{tabular} \\
    \midrule
    POPE & 88.84 & 89.11 \\
    \midrule
    Vanilla & 98.45 & 97.67 \\
    Visual Fixation & 80.62 & 74.03 \\
    False-Premise & 81.40 & 88.37 \\
    Premise-Induced Visual Fixation & 58.14 & 58.14 \\
    \bottomrule
  \end{tabular}
  }
  \caption{Reported POPE accuracy~\citep{huang2026many} compared with PVFH-Bench accuracy for Qwen3-VL models.}
  \label{tab:appendix_pope_pvfh_qwen3vl}
\end{table*}

\subsection{POPE Comparison for Qwen3-VL}
\label{sec:appendix_pope_comparison}

Table~\ref{tab:appendix_pope_pvfh_qwen3vl} contrasts reported accuracy with our four PVFH-Bench settings for the two Qwen3-VL variants used in our experiments.
POPE focuses on binary object-existence queries, while PVFH-Bench further targets visually confusable objects and false-premise spatial fixation.
The core Premise-Induced Visual Fixation Query setting is much harder than POPE for both models: Qwen3-VL-8B drops from 88.84 on POPE to 58.14 on Premise-Induced Visual Fixation Query, while Qwen3-VL-32B drops from 89.11 to 58.14.

\subsection{Model Licenses.}
All evaluated models are used in accordance with their respective licenses.
Qwen-VL models are released under the Tongyi Qianwen License;
Gemma models are released under the Gemma Terms of Use;
Seed models are accessed through publicly available API endpoints;
LLaMA models are released under the Meta LLaMA License.
We use these models solely for research evaluation purposes consistent with their intended use.

\subsection{Dataset License.}
Our benchmark is constructed from images in the MSCOCO 2014 validation set \citep{lin2014microsoft}, which is released under a Creative Commons Attribution 4.0 License (CC BY 4.0).
We do not modify the images and use them in compliance with the dataset license terms.
Our generated questions and annotations will be released under the same CC BY 4.0 license.

\subsection{Annotator Information and Ethical Considerations}
One human annotator participated in the data labeling process. 
The annotator held the bachelor’s degree and came from the academic background.
They volunteered his/her time, as part of an academic collaboration.
Prior to participation, the annotator provided informed consent for the use of the data in this study. 

\end{document}